\documentclass[10pt,twocolumn,letterpaper]{article}

\usepackage{wacv}      

\usepackage[accsupp]{axessibility}
\definecolor{wacvblue}{rgb}{0.21,0.49,0.74}
\usepackage[pagebackref,breaklinks,colorlinks,allcolors=wacvblue]{hyperref}

\def\wacvPaperID{691} 
\def\confName{WACV}
\def\confYear{2027}

\title{DGC-Net: A Depth-Geometric Conditioning Network for Projection-Supervised OCTA Segmentation}

\author{
\begin{tabular}{cccc}
Lucas Santiago$^{1,2}$ &
Bisheng Wang$^{1}$ &
Jaime S. Cardoso$^{1,2}$ &
Tom\'{a}s Ribeiro$^{1,2}$
\end{tabular}
\\[0.8em]
$^{1}$INESC TEC, Porto, Portugal\\
$^{2}$Faculty of Engineering, University of Porto (FEUP), Porto, Portugal
\\[0.5em]
{\tt\small up202104660@up.pt\quad bisheng.wang@inesctec.pt\quad jaime.cardoso@inesctec.pt\quad up202007710@up.pt}
}

\begin{document}
\maketitle
\begin{abstract}
OCTA segmentation is commonly formulated as a projection supervised task: models observe 3D volumetric scans but are supervised by 2D en-face semantic maps. Existing 2D and 3D-to-2D methods either discard axial information or compress depth responses into latent projections, leaving vessel crossings, diffuse capillary regions, and FAZ boundaries vulnerable to projection-induced ambiguity. We propose DGC-Net, a Depth-Geometric Conditioning Network that recasts 3D-to-2D OCTA segmentation as depth-geometrically conditioned prediction, where projected features are decoded with compact descriptors of axial vascular evidence. DGC-Net preserves local depth geometry through Depth-Moment Conditioning (DM), which injects axial localization and dispersion cues into the projection stream, and captures global depth layout through Holistic Depth-Geometry Modulation (HDM), which derives a raw-volume axial prior for final 2D decoding. 
Experiments on the OCTA-500 6mm and 3mm subsets show that DGC-Net achieves 83.58\% and 88.62\% mIoU, respectively, reaching state-of-the-art performance.
\end{abstract}    
\section{Introduction}

Optical Coherence Tomography Angiography (OCTA)~\cite{jia2012split,kashani2017optical} provides non-invasive volumetric observations of retinal vasculature, enabling detailed analysis of capillaries, arteries, veins, and the foveal avascular zone (FAZ). Accurate segmentation of these structures supports quantitative assessment of retinal diseases, as FAZ morphology, capillary dropout, and artery--vein abnormalities relate closely to diabetic retinopathy, hypertensive retinopathy, and glaucoma. In practice, OCTA vessel segmentation is typically formulated as an en-face task: the input is a 3D OCTA volume, while the available annotation and clinical output are 2D projection maps. This creates a projection-supervised setting, where the model must predict a 2D semantic layout while preserving the volumetric cues that remain discriminative after projection.

A common solution converts the OCTA volume into a 2D projection and applies standard 2D segmentation networks~\cite{unet,unetpp,attunet}. Although efficient, this collapses the axial dimension before representation learning, leaving the model reliant on projected intensity and texture alone. Recent 3D-to-2D methods alleviate this by learning projection operators or retaining height-wise responses: IPN and IPN-V2~\cite{ipn,octa500} progressively compress 3D features into 2D, while H2C-Net~\cite{h2cnet} converts the height dimension into channels. These methods demonstrate the value of volumetric information, but they mainly improve how depth responses are compressed or retained, rather than explicitly exposing the axial geometry that conditions the final semantic decision.

The core challenge is that 3D-to-2D projection is inherently many-to-one: distinct axial configurations---a localized vessel, a diffuse capillary response, or overlapping vessels at different depths---can share the same 2D appearance yet require different labels. This is acute for capillary--artery--vein--FAZ (CAVF) segmentation, where artery--vein crossings overlap in the en-face plane while separated along depth, and FAZ delineation depends on the broader organization and absence of vascular evidence. Once this axial geometry is absorbed into an opaque projected feature, a higher-capacity 2D decoder alone cannot recover the missing disambiguating information.

We therefore recast projection-supervised OCTA segmentation as depth-geometrically conditioned prediction: the decoder should receive not only the compressed en-face response but also compact descriptors of how vascular evidence is distributed along depth. To this end, we propose the \textbf{Depth-Geometric Conditioning Network (DGC-Net)}, which preserves two levels of depth geometry. The \emph{Depth-Moment Conditioning} (DM) module models local axial evidence as a depth-coordinate distribution and embeds its first two moments---axial localization and dispersion---into the projection stream, letting the network distinguish localized, diffuse, and overlapping responses before spatial collapse. The \emph{Holistic Depth-Geometry Modulation} (HDM) module extracts a raw-volume global axial prior to condition the final 2D representation with macroscopic depth-layout information that window-level statistics cannot capture.

In summary, our contributions are three-fold:
\begin{itemize}
\item We identify projection-induced depth ambiguity as a central challenge in projection-supervised OCTA segmentation and formulate depth-geometric conditioning as a principled solution.
\item We propose DGC-Net, integrating local Depth-Moment Conditioning (DM) and global Holistic Depth-Geometry Modulation (HDM) to preserve axial localization, dispersion, and global layout during 3D-to-2D projection.
\item We validate DGC-Net on OCTA-500 and DERMA-OCTA with per-volume significance testing, ablations, and depth-moment visualizations, achieving state-of-the-art or competitive results against strong 2D and 3D-to-2D baselines.
\end{itemize}
\section{Related Work}

\subsection{OCTA Datasets and Vessel Segmentation}

The development of automated OCTA segmentation has been closely tied to the availability of annotated benchmarks. Early datasets, such as FAZID~\cite{agarwal2020foveal} and ROSE~\cite{ma2021rose:}, mainly focus on 2D projection images and binary vessel or FAZ segmentation. The OCTA-500 benchmark~\cite{octa500} extends this setting by providing paired volumetric OCTA scans and multi-class en-face annotations for capillaries, arteries, veins, and the foveal avascular zone (FAZ). This benchmark introduces a more challenging projection-supervised setting, where models receive 3D observations but are supervised by 2D semantic labels. Beyond retinal imaging, DERMA-OCTA~\cite{Rotunno2025DERMAOCTA} provides volumetric OCTA scans with vessel annotations, enabling evaluation of OCTA segmentation methods across different tissue domains.

Most early deep learning methods for OCTA segmentation operate on 2D projection maps. U-Net~\cite{unet} and its variants~\cite{unetpp,attunet,resunet} establish strong baselines by learning hierarchical spatial features from en-face images. Subsequent methods introduce task-specific designs for vascular structures. For example, CS-Net~\cite{csnet} integrates spatial and channel attention for curvilinear vessel segmentation, while OCTA-Net~\cite{ma2021rose:} adopts a coarse-to-fine strategy to handle vessels at different scales. Transformer-based architectures, such as Swin-UNet~\cite{cao2022swin}, further improve global context modeling through self-attention. However, these 2D methods rely on fixed projections before learning. Once the axial dimension is collapsed into a 2D image, depth-dependent vascular organization is no longer explicitly available, making artery-vein crossings and overlapping vascular responses difficult to disambiguate.

\subsection{Volumetric Cues and 3D-to-2D Projection for OCTA}

To exploit volumetric OCTA data under 2D supervision, recent methods have studied learnable 3D-to-2D projection. IPN and IPN-V2~\cite{ipn,octa500} replace hand-crafted projections with Feature Projection Modules (FPMs), progressively compressing 3D volumes into 2D feature maps. H2C-Net~\cite{h2cnet} converts the axial dimension into channels and adopts a GPC-Net decoder to preserve height-wise responses, while LA-Net~\cite{Yangetal2024} uses layer-wise attention to suppress artifact-contaminated slices before projection. These methods demonstrate the value of volumetric cues, but primarily focus on compressing, rearranging, or selecting depth responses. In contrast, DGC-Net explicitly conditions semantic prediction on the axial geometry of vascular evidence.

Explicit depth cues have also been explored for OCTA segmentation. The Depth Channel Index Map (DCIM) estimates the axial centroid of vascular responses from the raw volume and has been used as an external pseudo-depth representation or self-supervised depth target~\cite{dcim}. Most closely related, the depth-enhanced cascaded framework of~\cite{dcim} combines this pseudo-depth cue with structure- and connectivity-aware losses, but keeps the depth cue external to the segmentation backbone. DGC-Net instead integrates depth geometry into the projection-supervised representation learning process by injecting DM and HDM modules, conditioning the projection operator itself rather than appending a precomputed depth map.

\section{Method}

\begin{figure*}[t]
    \centering
    \includegraphics[width=0.85\linewidth]{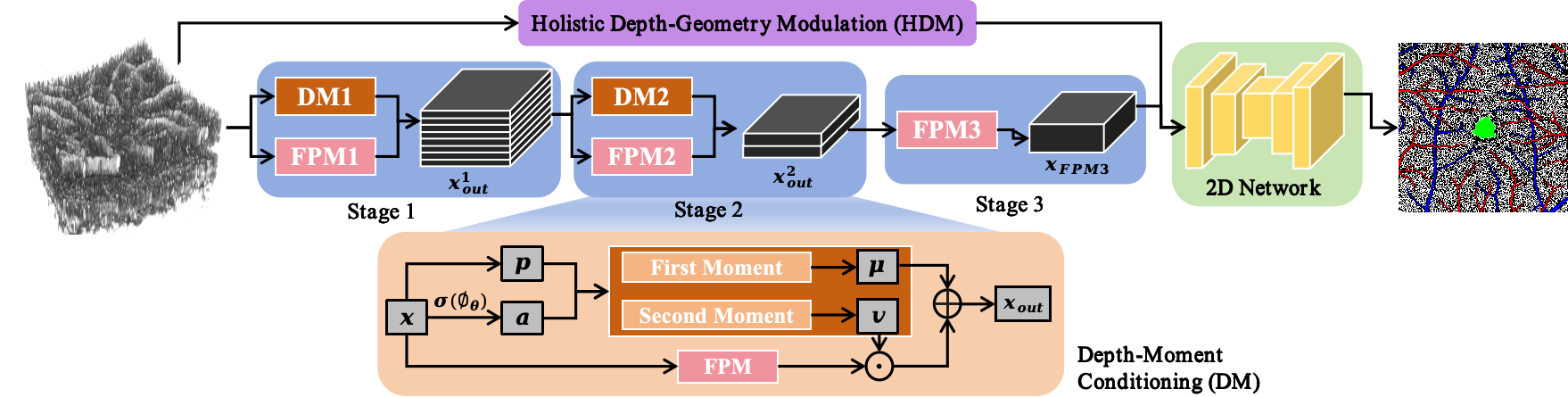}
    \vspace{-0.05in}
    \caption{\textbf{Architectural overview of DGC-Net.} Given a 3D OCTA volume, DGC-Net performs progressive 3D-to-2D projection through three projection stages. In the first two stages, the proposed Depth-Moment Conditioning (DM, bottom) module estimates local axial localization and dispersion cues, and fuses them with FPM features before further projection. In parallel, the Holistic Depth-Geometry Modulation (HDM) branch extracts a global depth-layout prior from the raw volume. The fused depth-geometric representation is finally decoded by a 2D segmentation network to produce the C/A/V/F segmentation map.}
    \vspace{-0.05in}
    \label{fig:framework}
\end{figure*}

\subsection{Depth-Geometric Conditioning under Projection Ambiguity}
\label{sec:DGC_Net}

We consider projection-supervised OCTA segmentation, where the input is a volumetric OCTA scan but the supervision is a 2D en-face semantic map. Given an OCTA volume $X \in \mathbb{R}^{1\times D\times H\times W}$ and a CAVF label map $Y \in \{1,\dots,K\}^{H\times W}$ defined only in the en-face plane---no volumetric vessel annotations exist---the goal is to predict capillaries, arteries, veins, and the FAZ from these 3D observations under 2D supervision. Existing 3D-to-2D OCTA networks can be generally written as
\begin{equation}
\hat{Y}=g_{\theta}(R), \qquad R=\mathcal{C}_{\theta}(X),
\end{equation}
where $\mathcal{C}_{\theta}$ projects the 3D volume into a 2D representation and $g_{\theta}$ denotes the 2D decoder.

The main difficulty is that $\mathcal{C}_{\theta}$ is a many-to-one projection. For an en-face location $u=(i,j)$, let $E_u=\{e_u(z)\}_{z=1}^{D}$ denote the latent axial vessel-evidence profile along the depth direction, and let $R_u$ be its projected response. Since distinct axial profiles may produce similar projected responses, $R_u$ corresponds to an approximate projection-equivalence class:
\begin{equation}
\Omega_{\delta}(R_u)=
\left\{
E_u \; \middle| \;
\left\|\Pi_{\theta}(E_u)-R_u\right\| \leq \delta
\right\},
\end{equation}
where $\Pi_{\theta}$ denotes the local projection effect at location $u$, and $\delta$ accounts for the tolerance of learned feature representations. Different profiles in $\Omega_{\delta}(R_u)$ may correspond to artery-vein crossings, diffuse capillary responses, or FAZ-related avascular patterns. Therefore, once axial organization is absorbed into an opaque 2D feature, a stronger decoder alone may not recover the missing depth cues.

To reduce this ambiguity, we formulate 3D-to-2D OCTA projection as depth-geometric conditioning. By depth geometry, we refer to the task-relevant axial organization of vascular evidence before projection. Let $G_u$ denote a compact descriptor of such depth geometry. Conditioning the decoder on both $R_u$ and $G_u$ refines the ambiguous set:
\begin{equation}
\Omega_{\delta}(R_u,G_u)=
\left\{
E_u \; \middle| \;
\left\|\Pi_{\theta}(E_u)-R_u\right\| \leq \delta,\;
\Gamma(E_u)=G_u
\right\},
\end{equation}
where $\Gamma(\cdot)$ extracts axial geometric information from the volumetric evidence. Since $\Omega_{\delta}(R_u,G_u) \subseteq \Omega_{\delta}(R_u)$, preserving task-relevant depth geometry can reduce the remaining semantic uncertainty:
\begin{equation}
H(Y_u \mid R_u,G_u) \leq H(Y_u \mid R_u),
\end{equation}
when $G_u$ captures information that is informative of $Y_u$.

Based on this principle, we propose the Depth-Geometric Conditioning Network (DGC-Net), as illustrated in Fig.~\ref{fig:framework}. DGC-Net instantiates $G_u$ through two complementary mechanisms. The Depth-Moment Conditioning (DM) module encodes local axial geometry from intermediate 3D features, preserving axial localization and dispersion during progressive projection. The Holistic Depth-Geometry Modulation (HDM) module extracts a global axial prior directly from the raw OCTA volume and modulates the final 2D representation with macroscopic depth-layout information. Together, DM and HDM transform 3D-to-2D OCTA segmentation from opaque feature compression into depth-geometrically conditioned semantic prediction under 2D supervision.

\subsection{Local Depth-Moment Conditioning (DM)}

The Depth-Moment Conditioning (DM) module instantiates the local component of depth-geometric conditioning. Given an intermediate 3D feature volume $\mathbf{x}^{(l)} \in \mathbb{R}^{C_l\times D_l\times H_l\times W_l}$ at projection stage $l$, where $C_l, D_l, H_l, W_l$ are the channel, depth, and spatial dimensions produced by the preceding FPM (with $C_0{=}1$ and $D_0{=}D$ at the raw input), a conventional Feature Projection Module (FPM)~\cite{octa500} produces a compressed representation
\begin{equation}
\mathbf{x}^{(l)}_{\mathrm{FPM}}=\Pi_l(\mathbf{x}^{(l)}).
\end{equation}
While this representation encodes projected appearance, it may hide the axial organization of local vascular evidence. DM therefore augments each projection stage with a compact moment embedding of the local depth distribution before full spatial collapse.

Specifically, we partition the depth axis into local windows aligned with the progressive FPM grouping. For an en-face location $u$ and a depth slice $d$ within a local window, let $p_d \in [0,1]$ denote the normalized axial coordinate. DM assigns a non-competitive evidence score to each depth slice:
\begin{equation}
a^{(l)}_{u,d}
=
\sigma
\left(
\phi_\theta
\left(
\mathbf{x}^{(l)}_{u,d}
\right)
\right),
\end{equation}
where $\sigma(\cdot)$ is the sigmoid and $\phi_\theta$ is a lightweight ($1{\times}1$ convolutional) scoring function. The sigmoid scoring treats axial evidence---the learned per-slice vascular response, not a binary mask---as a multi-response measure rather than a single normalized mode, which is suitable for OCTA because multiple vessels, or no vessel, may appear along the same en-face ray.

We then summarize this local axial evidence distribution by its first and second central moments:
\begin{equation}
\begin{aligned}
\mu^{(l)}_u
&=
\mathbb{E}_{\alpha^{(l)}_u}[p]
=
\frac{
\sum_{d=1}^{S_l}
a^{(l)}_{u,d} p_d
}{
\sum_{d=1}^{S_l}
a^{(l)}_{u,d}
+\epsilon
}, \\
\nu^{(l)}_u
&=
\mathbb{E}_{\alpha^{(l)}_u}
\left[
\left(
p-\mu^{(l)}_u
\right)^2
\right]
=
\frac{
\sum_{d=1}^{S_l}
a^{(l)}_{u,d}
\left(
p_d-\mu^{(l)}_u
\right)^2
}{
\sum_{d=1}^{S_l}
a^{(l)}_{u,d}
+\epsilon
}.
\end{aligned}
\end{equation}
Here, $\mu^{(l)}_u$ encodes axial localization, indicating where the local vascular evidence is concentrated, while $\nu^{(l)}_u$ encodes axial dispersion, reflecting whether the evidence is sharp, thick, diffuse, or distributed across multiple layers. Together, $(\mu^{(l)}_u,\nu^{(l)}_u)$ forms a moment embedding of local depth geometry.

Finally, DM injects this moment embedding into the projected feature:
\begin{equation}
\mathbf{x}^{(l)}_{\mathrm{out}}
=
\left[
\psi_\mu
\left(
\mu^{(l)}
\right),
\;
\mathbf{x}^{(l)}_{\mathrm{FPM}}
\odot
\left(
1+
\psi_\nu
\left(
\nu^{(l)}
\right)
\right)
\right],
\end{equation}
where $\psi_\mu$ and $\psi_\nu$ lift the moment maps into channel-compatible feature maps. Thus, $\mu^{(l)}$ provides explicit axial coordinate cues, while $\nu^{(l)}$ adaptively modulates regions with greater depth ambiguity. This multiplicative modulation is related to feature-wise linear modulation (FiLM)~\cite{perez2018film}, but differs in its source: rather than predicting modulation parameters from an external conditioning input, DM derives the gate from the intrinsic axial moment statistics of the feature volume itself. Applied hierarchically, DM carries local depth geometry through the 3D-to-2D projection stream before the axial dimension is fully collapsed.

\subsection{Holistic Depth-Geometry Modulation (HDM)}

While DM preserves local depth geometry within intermediate projection windows, it does not directly provide a volume-level axial layout prior. To complement this local conditioning, HDM introduces a geometry-initialized global conditioning branch. Instead of learning global depth attention only from compressed features, HDM first derives a stable axial centroid prior from the raw OCTA volume and then learns to align this prior with the semantic feature space of the projection stream.

For each en-face coordinate $u=(i,j)$, HDM computes a confidence-weighted axial centroid:
\begin{equation}
\mathbf{M}(i,j)
=
\frac{
\sum_{d=1}^{D}
p_d \cdot \mathbf{X}(d,i,j)^2
}{
\sum_{d=1}^{D}
\mathbf{X}(d,i,j)^2
+\epsilon
},
\end{equation}
where $p_d \in [0,1]$ is the normalized depth coordinate. The squared intensity acts as confidence weighting, suppressing weak speckle-like responses while emphasizing high-confidence vascular evidence. Thus, $\mathbf{M}$ provides a compact raw-volume descriptor of global axial layout.

HDM then transforms this geometric prior into a learnable conditioning feature and fuses it with the final projected representation:
\begin{equation}
\mathbf{x}_{\mathrm{UNet}}
=
\left[
\mathbf{x}_{\mathrm{FPM3}},
\psi_{\mathrm{H}}
\left(
\mathbf{M}
\right)
\right],
\end{equation}
where $\mathbf{x}_{\mathrm{FPM3}}$ is the final-stage projected 2D feature (Fig.~\ref{fig:framework}) and $\psi_{\mathrm{H}}$ is a lightweight convolutional transformation that aligns the centroid prior with the decoder feature space. The centroid $\mathbf{M}$ follows the Depth Channel Index Map (DCIM) formulation introduced in~\cite{dcim}. Our contribution is not the centroid itself but how it is used: whereas~\cite{dcim} employs the DCIM as an external pseudo-depth input alongside the projected image, HDM embeds it inside the 3D-to-2D network as an internal, learnable global conditioning branch, complementing DM with macroscopic depth-layout information after local progressive projection.

\subsection{Training objective.} 
\label{sec:gce}

We train DGC-Net with a combination of pixel-wise Generalised Cross-Entropy (GCE)~\cite{gce} and Dice loss. GCE is used as a robust alternative to standard cross-entropy for dense projection-supervised segmentation, where some boundary and capillary pixels may provide less reliable hard supervision after 3D-to-2D projection. For a valid pixel $u$ with ground-truth class $y_u$, the GCE term is computed as
\begin{equation}
\mathcal{L}_{\mathrm{GCE}}
=
\frac{1}{|\Omega|}
\sum_{u \in \Omega}
\frac{1-p_{u,y_u}^{q}}{q},
\end{equation}
where $\Omega$ is the set of valid labeled pixels, $p_{u,y_u}$ is the predicted probability of class $y_u$, and $q \in (0,1]$ controls the robustness of the loss. The final objective is
\begin{equation}
\mathcal{L}_{\mathrm{total}}
=
\lambda_{\mathrm{gce}}
\mathcal{L}_{\mathrm{GCE}}
+
\mathcal{L}_{\mathrm{Dice}}.
\end{equation}
The Dice term provides standard overlap supervision and helps mitigate class imbalance. We treat this objective as a training choice complementary to the proposed depth-geometric conditioning architecture.
\section{Experiments}

\subsection{Datasets and Evaluation Metrics}

\textbf{OCTA-500}~\cite{octa500} serves as our primary benchmark. It contains 500 volumetric OCTA scans with raw 3D volumes and pixel-level en-face annotations for capillaries (C), arteries (A), veins (V), and the foveal avascular zone (FAZ). We evaluate both standard field-of-view subsets. The \textbf{6mm} subset includes 300 volumes of size $128 \times 400 \times 400$, split into 240/10/50 subjects for training/validation/testing (IDs 10001--10240/10241--10250/10251--10300). The \textbf{3mm} subset includes 200 volumes of size $128 \times 304 \times 304$, split into 140/10/50 subjects (IDs 10301--10440/10441--10450/10451--10500), strictly adhering to the standard benchmark protocol.

\textbf{DERMA-OCTA}~\cite{Rotunno2025DERMAOCTA} is utilized as an independent, out-of-domain dataset to rigorously validate the architectural generalizability of our framework. It comprises 330 dermatological OCTA volumes ($90 \times 512 \times 483$ voxels) from 74 subjects presenting with various skin conditions. Following the official split, we use 199/68/45 volumes for training/validation/testing and evaluate full-depth 2D binary vessel segmentation.

\textbf{Evaluation Metrics.} To assess segmentation performance, we employ mean Dice Similarity Coefficient (mDice) and mean Intersection over Union (mIoU). Both metrics are calculated per class and averaged to yield the final scores. To assess whether performance differences are statistically significant, we apply per-volume paired Wilcoxon signed-rank tests on the three-seed-averaged mean IoU across the 50 test volumes.

\subsection{Implementation Details}

DGC-Net is implemented in PyTorch and trained for 200 epochs with a batch size of 2. We use the Adam optimizer with a tri-stage learning rate schedule. The learning rate is linearly warmed up from $5 \times 10^{-5}$ to $5 \times 10^{-4}$ during the first 30 epochs, kept constant for the next 50 epochs, and exponentially decayed back to $5 \times 10^{-5}$ over the remaining 120 epochs. Standard geometric augmentations, including random horizontal flipping, vertical flipping, and $90^\circ$ rotations, are applied during training with a probability of 0.25.

Unless otherwise specified, the final DGC-Net adopts GPC-Net as the terminal 2D decoder, following H2C-Net~\cite{h2cnet}, to ensure a fair comparison with the strongest existing 3D-to-2D OCTA segmentation baseline. The final model is trained with the robust objective described in Sec.~\ref{sec:gce}: we first optimize the network with CE and Dice loss for 20 epochs to stabilize early convergence, and then replace the CE term with GCE using $q=0.5$ for the remaining epochs.

For the ablation study, we use a U-Net decoder and CE-Dice supervision by default, so that the effects of the proposed depth-geometric conditioning modules can be isolated from a stronger 2D backend and the robust training objective. The GCE term is added in the final ablation setting to evaluate its auxiliary contribution. When adapting DGC-Net to the DERMA-OCTA dataset, the depth dimension of each raw volume is zero-padded from 90 to 96 slices to satisfy the divisibility required by the hierarchical FPM grouping schedule. All experiments are conducted on a single NVIDIA A100 GPU.

\subsection{Ablation Study}

We ablate on OCTA-6mm, where the larger field of view makes projection ambiguity most pronounced. Unless noted, all variants share the IPNv2 backbone, U-Net decoder, and CE-Dice supervision; GPC-Net and GCE-Dice appear only in the final configuration. We report mean $\pm$ std over three runs and discuss means in the text.

\begin{table*}[t]
\centering
\setlength{\tabcolsep}{3pt}
\renewcommand{\arraystretch}{0.95}
\resizebox{0.9\linewidth}{!}{
\begin{tabular}{lcccccc}
\hline
Method & C & A & V & F & mIoU & mDice \\
\hline
IPNv2 (baseline)
& $0.8432{\pm}0.0040$ 
& $0.7704{\pm}0.0017$ 
& $0.7806{\pm}0.0019$ 
& $0.8761{\pm}0.0004$ 
& $0.8176{\pm}0.0010$ 
& $0.8965{\pm}0.0007$ \\

- DM ($\mu$ only, softmax)
& $0.8523{\pm}0.0013$ 
& $0.7746{\pm}0.0028$ 
& $0.7854{\pm}0.0012$ 
& $0.8779{\pm}0.0047$ 
& $0.8225{\pm}0.0001$ 
& $0.8996{\pm}0.0001$ \\

- DM ($\mu$ only, sigmoid)
& $0.8500{\pm}0.0020$ 
& $0.7766{\pm}0.0029$ 
& $0.7856{\pm}0.0009$ 
& $0.8848{\pm}0.0016$ 
& $0.8242{\pm}0.0010$ 
& $0.9009{\pm}0.0008$ \\

- DM ($\nu$ only, sigmoid)
& $0.8495{\pm}0.0016$
& $0.7749{\pm}0.0012$
& $0.7833{\pm}0.0014$
& $\textbf{0.8859}{\pm}\textbf{0.0041}$
& $0.8234{\pm}0.0012$
& $0.9006{\pm}0.0008$ \\

- DM ($\mu+\nu$, sigmoid)
& $\textbf{0.8543}{\pm}\textbf{0.0015}$ 
& $\textbf{0.7790}{\pm}\textbf{0.0017}$ 
& $\textbf{0.7880}{\pm}\textbf{0.0032}$ 
& $0.8818{\pm}0.0048$ 
& $\textbf{0.8258}{\pm}\textbf{0.0001}$ 
& $\textbf{0.9017}{\pm}\textbf{0.0001}$ \\
\hline
\end{tabular}
}
\vspace{-0.05in}
\caption{Ablation of local depth-moment conditioning on OCTA-6mm. The $\mu$-only variants inject axial localization without the dispersion gate, while the $\nu$-only variant removes the localization feature and keeps only dispersion-guided modulation. The $\mu+\nu$ setting denotes the default DM configuration. C: Capillary, A: Artery, V: Vein, F: FAZ.}
\vspace{-0.05in}
\label{tab:moment_ablation}
\end{table*}

\textbf{Effect of local depth-moment conditioning.}
Table~\ref{tab:moment_ablation} evaluates the DM design choices. From the IPNv2 baseline, $\mu$-only conditioning with softmax scoring improves mIoU $0.8176{\to}0.8225$, showing that explicit axial localization already helps beyond opaque projection. Replacing softmax with sigmoid scoring further improves the mIoU to $0.8242$, with a clear gain in the FAZ class from $0.8779$ to $0.8848$. This trend is consistent with our formulation of axial vessel evidence as a non-competitive measure: OCTA responses along the same en-face ray are not necessarily single-mode, and avascular regions may contain no dominant vessel response.

Adding the dispersion-guided $\nu$-gate to the sigmoid-based $\mu$ conditioning further increases the mIoU to $0.8258$. The gains are mainly observed in capillaries, arteries, and veins, indicating that axial dispersion is useful for modeling thick, diffuse, or overlapping vascular evidence. Meanwhile, the FAZ score slightly decreases compared with the $\mu$-only sigmoid variant, which suggests that local dispersion alone does not fully capture broader avascular layout. This observation motivates the use of HDM as a complementary global depth-geometric condition.

The dispersion gate is also effective in isolation: the $\nu$-only variant improves the baseline mIoU from $0.8176$ to $0.8234$ and, notably, attains the highest FAZ IoU of all DM variants ($0.8859$). This indicates that axial dispersion alone is a strong cue for the FAZ, whose defining property is the diffuse, low-concentration axial profile of the surrounding capillary bed rather than a sharply localized vessel response. On the vessel classes, however, $\nu$-only trails the combined $\mu{+}\nu$ configuration, confirming that dispersion is complementary to---but not a substitute for---explicit axial localization.

\begin{table*}[t]
\centering
\setlength{\tabcolsep}{3pt}
\renewcommand{\arraystretch}{0.95}
\resizebox{\linewidth}{!}{
\begin{tabular}{lcccccc}
\hline
Method & C & A & V & F & mIoU & mDice \\
\hline
IPNv2 (baseline) 
& $0.8432{\pm}0.0040$ 
& $0.7704{\pm}0.0017$ 
& $0.7806{\pm}0.0019$ 
& $0.8761{\pm}0.0004$ 
& $0.8176{\pm}0.0010$ 
& $0.8965{\pm}0.0007$ \\

+ DM
& $0.8543{\pm}0.0015$ 
& $0.7790{\pm}0.0017$ 
& $0.7880{\pm}0.0032$ 
& $0.8818{\pm}0.0048$ 
& $0.8258{\pm}0.0001$ 
& $0.9017{\pm}0.0001$ \\

+ HDM
& $0.8521{\pm}0.0011$
& $0.7793{\pm}0.0021$
& $0.7904{\pm}0.0020$
& $0.8858{\pm}0.0012$
& $0.8269{\pm}0.0010$
& $0.9026{\pm}0.0006$ \\

+ DM + HDM 
& $0.8561{\pm}0.0015$ 
& $0.7809{\pm}0.0031$ 
& $0.7905{\pm}0.0037$ 
& $0.8945{\pm}0.0050$ 
& $0.8305{\pm}0.0010$ 
& $0.9049{\pm}0.0006$ \\

+ DM + HDM + GCE 
& $0.8571{\pm}0.0012$ 
& $0.7818{\pm}0.0008$ 
& $0.7938{\pm}0.0019$ 
& $\mathbf{0.8983{\pm}0.0068}$ 
& $0.8328{\pm}0.0012$ 
& $0.9063{\pm}0.0011$ \\

\hline
+ DM + HDM + GCE + GPC-Net 
& $\mathbf{0.8592{\pm}0.0025}$ 
& $\mathbf{0.7901{\pm}0.0015}$ 
& $\mathbf{0.8011{\pm}0.0010}$ 
& $0.8930{\pm}0.0020$ 
& $\mathbf{0.8358{\pm}0.0001}$ 
& $\mathbf{0.9083{\pm}0.0001}$ \\
\hline
\end{tabular}
}
\vspace{-0.05in}
\caption{Component-wise ablation study on the OCTA-6mm benchmark. Unless otherwise specified, all variants use a U-Net decoder and CE-Dice supervision. The final row adopts GPC-Net as the decoder and GCE-Dice training for comparison with H2C-Net. DM denotes the default sigmoid-based $\mu+\nu$ configuration. C: Capillary, A: Artery, V: Vein, F: FAZ.}
\vspace{-0.05in}
\label{tab:ablation}
\end{table*}

\textbf{Complementarity between local and global depth geometry.}
Table~\ref{tab:ablation} evaluates local and global depth-geometric conditioning. DM improves the IPNv2 baseline from $0.8176$ to $0.8258$ mIoU, while HDM alone reaches $0.8269$, showing that both local depth moments and raw-volume axial layout provide useful cues beyond opaque projection. Their gains are partially overlapping, as both describe axial vascular organization, but they operate at different scales: DM conditions intermediate projection stages with local localization and dispersion, whereas HDM supplies a global axial prior from the raw volume. Combining them further improves mIoU to $0.8305$, with the largest gain on FAZ ($0.8818/0.8858 \rightarrow 0.8945$). This suggests that local vessel classes already benefit from DM, while FAZ segmentation particularly benefits from the macroscopic depth-layout condition provided by HDM.

\textbf{Effect of the robust objective and decoder choice.}
Replacing CE with GCE after the warm-up stage further improves the mIoU from $0.8305$ to $0.8328$. The improvement is moderate but consistent in the mean performance, suggesting that the robust objective can reduce the influence of less reliable hard labels in projection-supervised training. We treat this loss as an auxiliary training choice rather than an independent architectural contribution.

Finally, using GPC-Net as the terminal decoder increases the mIoU to $0.8358$ and the mDice to $0.9083$. Since H2C-Net also adopts GPC-Net, this setting is used as our final configuration for fair comparison with the strongest 3D-to-2D OCTA segmentation baseline. Importantly, the component analysis above is conducted before introducing GPC-Net, so that the contribution of DM and HDM is not confounded with the use of a stronger 2D decoder.

\begin{figure}[t]
\centering
\includegraphics[width=0.9\linewidth]{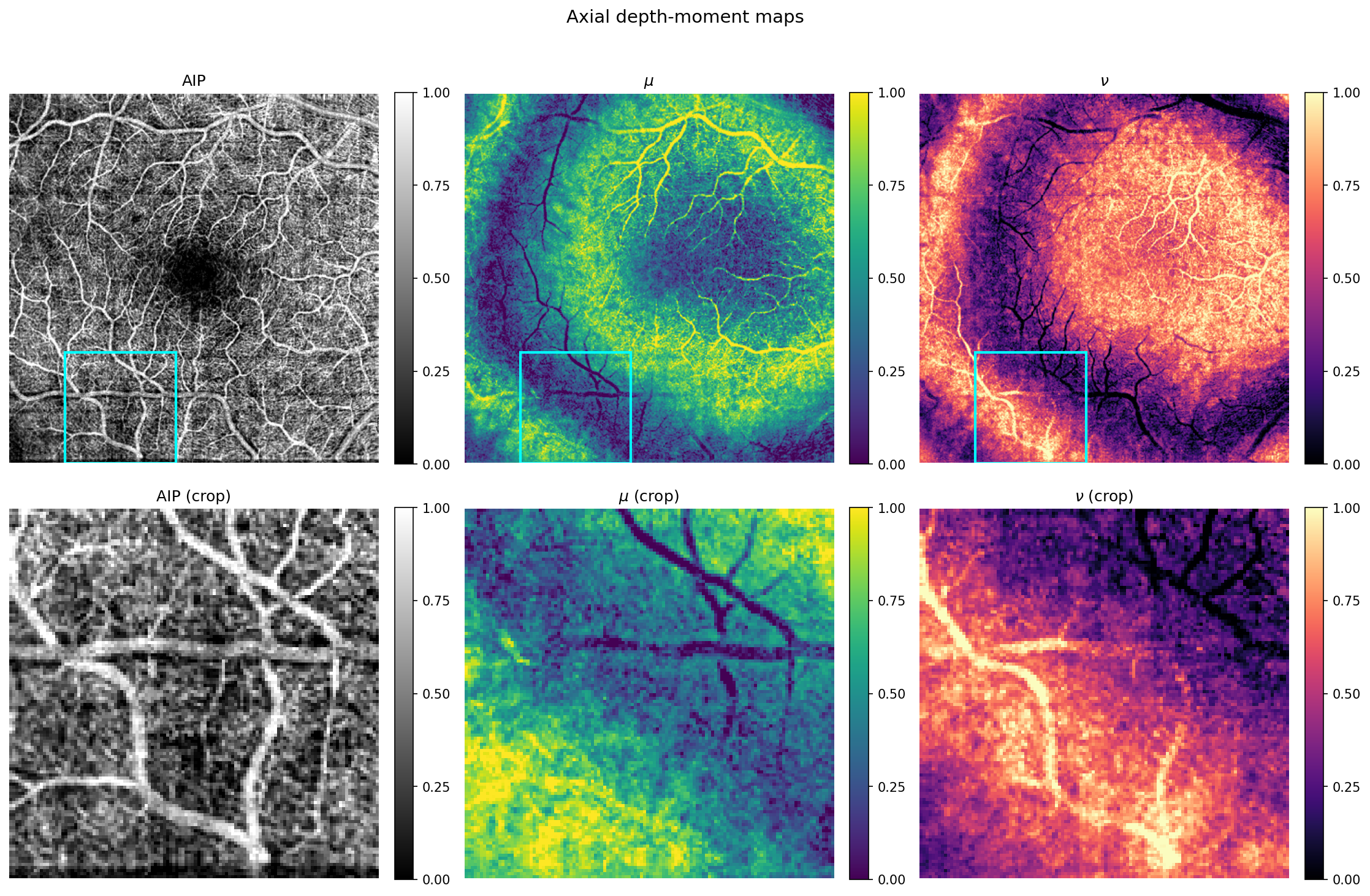} 
\vspace{-0.1in}
\caption{Axial depth-moment maps for a representative OCTA-6mm test subject. Columns: en-face AIP, axial localization $\mu$, axial dispersion $\nu$. Top row: full maps (cyan box marks the crop region); bottom row: zoomed crop at a vascular junction.}
\vspace{-0.05in}
\label{fig:moments}
\end{figure}

\textbf{Visualization of depth-moment maps.}
Figure~\ref{fig:moments} visualizes the axial localization map $\mu$ and axial dispersion map $\nu$ from the first DM module on a representative 6mm test subject. Both descriptors exhibit strongly vessel-aligned spatial structure. The localization map $\mu$ assigns coherent axial positions along individual vessels, tracing the vascular tree against a smoothly varying background. The dispersion map $\nu$ is low along well-defined vessels, where axial evidence concentrates at a single depth, and elevated in perifoveal and capillary-dense regions, where vascular evidence is spread across multiple depths---precisely the configurations most affected by projection-induced ambiguity. The zoomed crop highlights a vascular junction, where $\nu$ responds at the overlap of structures occupying distinct depth strata. These observations confirm that the DM module recovers interpretable depth geometry before projection, consistent with the accuracy gains in Tables~\ref{tab:moment_ablation} and~\ref{tab:ablation}.

\begin{table*}[t]
\setlength{\tabcolsep}{3pt}
\centering
\setlength{\tabcolsep}{3pt}
\renewcommand{\arraystretch}{0.95}
\resizebox{\textwidth}{!}{
\begin{tabular}{lrrrcccccc}
\hline
Method & Params(M) & FLOPs(G) & Speed(ms) & C & A & V & F & mIoU & mDice \\
\hline
\multicolumn{10}{c}{\textit{2D Methods}} \\
UNet~\cite{unet} 
& 1.377 & 106.450 & 4.760
& $0.7103{\pm}0.0012$
& $0.7703{\pm}0.0041$
& $0.7793{\pm}0.0036$
& $0.8766{\pm}0.0014$
& $0.7841{\pm}0.0023$
& $0.8756{\pm}0.0015$ \\

UNet++~\cite{unetpp}
& 3.023 & 373.079 & 16.432
& $0.7126{\pm}0.0008$
& $0.7703{\pm}0.0018$
& $0.7810{\pm}0.0008$
& $0.8799{\pm}0.0033$
& $0.7860{\pm}0.0001$
& $0.8766{\pm}0.0003$ \\

UNet3+~\cite{huang2020unet}
& 4.036 & 478.863 & 15.108
& $0.7112{\pm}0.0010$
& $0.7689{\pm}0.0024$
& $0.7794{\pm}0.0009$
& $0.8723{\pm}0.0070$
& $0.7829{\pm}0.0017$
& $0.8748{\pm}0.0012$ \\

Attention UNet~\cite{attunet}
& 1.405 & 153.268 & 7.357
& $0.7135{\pm}0.0018$
& $0.7721{\pm}0.0016$
& $0.7793{\pm}0.0036$
& $0.8747{\pm}0.0048$
& $0.7849{\pm}0.0017$
& $0.8756{\pm}0.0012$ \\

UNet-FPN~\cite{dcim}
& 2.790 & 200.643 & 9.146
& $0.7107{\pm}0.0028$
& $0.7794{\pm}0.0028$
& $0.7888{\pm}0.0013$
& $0.8816{\pm}0.0110$
& $0.7901{\pm}0.0023$
& $0.8788{\pm}0.0024$ \\

\hline
\multicolumn{10}{c}{\textit{3D-to-2D Methods}} \\
IPNv2~\cite{octa500}
& 3.871 & 594.279 & 22.727
& $0.8432{\pm}0.0040$
& $0.7704{\pm}0.0017$
& $0.7806{\pm}0.0019$
& $0.8761{\pm}0.0004$
& $0.8176{\pm}0.0010$
& $0.8965{\pm}0.0007$ \\

H2C-Net~\cite{h2cnet}
& 19.978 & 2244.353 & 87.509
& $\textbf{0.8664}{\pm}\textbf{0.0020}$
& $0.7862{\pm}0.0032$
& $0.7950{\pm}0.0026$
& $0.8768{\pm}0.0033$
& $0.8311{\pm}0.0014$
& $0.9049{\pm}0.0011$ \\

\textbf{DGC-Net (ours)}
& 5.482 & 860.324 & 66.555
& $0.8592{\pm}0.0025$
& $\textbf{0.7901}{\pm}\textbf{0.0015}$
& $\textbf{0.8011}{\pm}\textbf{0.0010}$
& $\textbf{0.8930}{\pm}\textbf{0.0020}$
& $\textbf{0.8358}{\pm}\textbf{0.0001}$
& $\textbf{0.9083}{\pm}\textbf{0.0001}$ \\
\hline
\end{tabular}
}
\vspace{-0.05in}
\caption{Comparison with state-of-the-art methods on the OCTA-6mm benchmark. Results are reported as mean $\pm$ standard deviation over three independent training runs. Params, FLOPs, and Speed denote model parameters, computational cost, and inference latency, respectively. C: Capillary, A: Artery, V: Vein, F: FAZ.}
\label{tab:sota_6mm}
\end{table*}

\begin{table*}[t]
\centering
\setlength{\tabcolsep}{3pt}
\renewcommand{\arraystretch}{0.95}
\resizebox{0.9\textwidth}{!}{
\begin{tabular}{lcccccc}
\hline
Method & C & A & V & F & mIoU & mDice \\
\hline
\multicolumn{7}{c}{\textit{2D Methods}} \\
UNet~\cite{unet} 
& $0.7854{\pm}0.0003$
& $0.8227{\pm}0.0011$
& $0.8103{\pm}0.0027$
& $0.9567{\pm}0.0012$
& $0.8438{\pm}0.0013$
& $0.9133{\pm}0.0007$ \\

UNet++~\cite{unetpp}
& $0.7865{\pm}0.0012$
& $0.8236{\pm}0.0013$
& $0.8112{\pm}0.0034$
& $0.9596{\pm}0.0011$
& $0.8452{\pm}0.0011$
& $0.9141{\pm}0.0007$ \\

UNet3+~\cite{huang2020unet}
& $0.7866{\pm}0.0007$
& $0.8199{\pm}0.0042$
& $0.8067{\pm}0.0023$
& $0.9605{\pm}0.0028$
& $0.8434{\pm}0.0012$
& $0.9130{\pm}0.0007$ \\

Attention UNet~\cite{attunet}
& $0.7872{\pm}0.0005$
& $0.8246{\pm}0.0011$
& $0.8129{\pm}0.0024$
& $0.9632{\pm}0.0013$
& $0.8470{\pm}0.0006$
& $0.9152{\pm}0.0004$ \\

UNet-FPN
& $0.7873{\pm}0.0008$
& $0.8346{\pm}0.0044$
& $0.8249{\pm}0.0052$
& $0.9620{\pm}0.0010$
& $0.8522{\pm}0.0025$
& $0.9184{\pm}0.0015$ \\

\hline
\multicolumn{7}{c}{\textit{3D-to-2D Methods}} \\
IPNv2~\cite{octa500}
& $0.8812{\pm}0.0021$
& $0.8307{\pm}0.0011$
& $0.8208{\pm}0.0024$
& $0.9591{\pm}0.0021$
& $0.8730{\pm}0.0006$
& $0.9308{\pm}0.0004$ \\

H2C-Net~\cite{h2cnet}
& $\textbf{0.9017}{\pm}\textbf{0.0015}$
& $0.8460{\pm}0.0018$
& $0.8353{\pm}0.0012$
& $0.9563{\pm}0.0023$
& $0.8848{\pm}0.0015$
& $0.9377{\pm}0.0009$ \\

\textbf{DGC-Net (ours)}
& $0.8935{\pm}0.0008$
& $\textbf{0.8480}{\pm}\textbf{0.0017}$
& $\textbf{0.8399}{\pm}\textbf{0.0023}$
& $\textbf{0.9633}{\pm}\textbf{0.0001}$
& $\textbf{0.8862}{\pm}\textbf{0.0006}$
& $\textbf{0.9385}{\pm}\textbf{0.0003}$ \\
\hline
\end{tabular}
}
\vspace{-0.05in}
\caption{Comparison with state-of-the-art methods on the OCTA-3mm benchmark. Results are reported as mean $\pm$ standard deviation over three independent training runs. C: Capillary, A: Artery, V: Vein, F: FAZ.}
\label{tab:sota_3mm}
\end{table*}

\subsection{Comparison with State-of-the-Art Methods}

To evaluate DGC-Net, we compare it with representative 2D and recent 3D-to-2D OCTA segmentation methods on the OCTA-500 benchmark. Results on the 6mm and 3mm subsets are reported in Tables~\ref{tab:sota_6mm} and~\ref{tab:sota_3mm}, respectively. For methods evaluated in our framework, we report mean $\pm$ standard deviation over three independent training runs.

\textbf{Results on the OCTA-6mm Benchmark.}
As shown in Table~\ref{tab:sota_6mm}, DGC-Net achieves $0.8358$ mIoU and $0.9083$ mDice on the OCTA-6mm subset. Compared with the strongest 2D baseline in our evaluation, UNet-FPN, DGC-Net improves the mIoU by 4.57 pp ($0.8358$ vs. $0.7901$). This gap reflects the core limitation of 2D pipelines: collapsing the volume to a fixed projection before learning discards the axial organization needed to disambiguate overlapping, depth-dependent structures.

Compared with the IPNv2 projection baseline, DGC-Net improves the mIoU from $0.8176$ to $0.8358$, with consistent gains across all four categories. The improvements are particularly clear for arteries, veins, and the FAZ, where DGC-Net increases the IoU by $+1.97$ pp, $+2.05$ pp, and $+1.69$ pp, respectively. These class-level gains are consistent with our motivation: local depth-moment conditioning provides axial localization and dispersion cues for vessel structures, while holistic depth-geometry modulation provides a global axial layout prior for macroscopic regions such as the FAZ.

Compared with H2C-Net, the strongest existing 3D-to-2D baseline, DGC-Net achieves higher mIoU ($0.8358$ vs.\ $0.8311$) and mDice ($0.9083$ vs.\ $0.9049$). A per-volume paired Wilcoxon signed-rank test confirms this improvement is statistically significant ($p = 6.3\times10^{-5}$, $n=50$; DGC-Net superior on $37/50$ volumes). The two methods take opposite approaches to depth: H2C-Net preserves the full height dimension by converting it into channels, relying on decoder capacity to disentangle the stacked responses, whereas DGC-Net explicitly conditions the projection on compact depth-geometric descriptors. Notably, DGC-Net attains this improvement with far fewer parameters ($5.48$M vs.\ $19.98$M) and lower computational cost, indicating that explicit geometric conditioning is a more parameter-efficient route to resolving projection ambiguity than increasing decoder capacity. The class-level pattern is consistent with this: H2C-Net retains a small advantage on capillaries, whose dense texture benefits from its high-capacity channel representation, while DGC-Net leads on arteries, veins, and the FAZ---the structures where axial localization and global depth layout matter most.

\textbf{Results on the OCTA-3mm Benchmark.} 
Table~\ref{tab:sota_3mm} reports results on the OCTA-3mm subset. DGC-Net attains the best overall performance ($0.8862$ mIoU, $0.9385$ mDice), improving over the IPNv2 baseline by $1.32$ pp mIoU and leading all 2D baselines by a wide margin. Relative to H2C-Net, DGC-Net achieves a higher mean mIoU ($0.8862$ vs.\ $0.8848$), but the margin is small: a per-volume paired Wilcoxon signed-rank test finds no significant difference between the two methods on 3mm ($p = 0.18$, $n=50$), in contrast to the significant margin on 6mm ($p = 6.3\times10^{-5}$). The methods are therefore statistically comparable at this resolution, with DGC-Net retaining consistent advantages on arteries, veins, and the FAZ. This is consistent with our expectation: the 3mm subset has a smaller field of view, reducing large-scale projection ambiguity and narrowing the gap between explicit depth conditioning and capacity-based alternatives. The advantage of depth-geometric conditioning is thus most pronounced on 6mm scans, where axial conflation is most severe, while remaining competitive on 3mm.

\textbf{Parameter Efficiency and Computational Cost.}
We report parameters, FLOPs, and inference speed on the OCTA-6mm setting in Table~\ref{tab:sota_6mm}. All measurements use a single NVIDIA A100 GPU at the OCTA-6mm resolution: 2D models take a $3 \times 400 \times 400$ input, while 3D-to-2D models take the full $128 \times 400 \times 400$ volume. Compared with H2C-Net, DGC-Net uses substantially fewer parameters ($5.48$M vs. $19.98$M), lower FLOPs ($860$G vs. $2244$G), and faster inference ($66.6$ ms vs. $87.5$ ms), while achieving higher mIoU ($0.8358$ vs. $0.8311$). Compared with IPNv2, DGC-Net introduces additional depth-geometric conditioning and a stronger 2D decoder, increasing computation but substantially improving segmentation accuracy ($0.8358$ vs. $0.8176$ mIoU). These results show that DGC-Net improves accuracy with a favorable efficiency trade-off among 3D-to-2D OCTA segmentation networks.

\subsection{Qualitative Results}

\begin{figure}[t]
\centering
\includegraphics[width=0.9\linewidth]{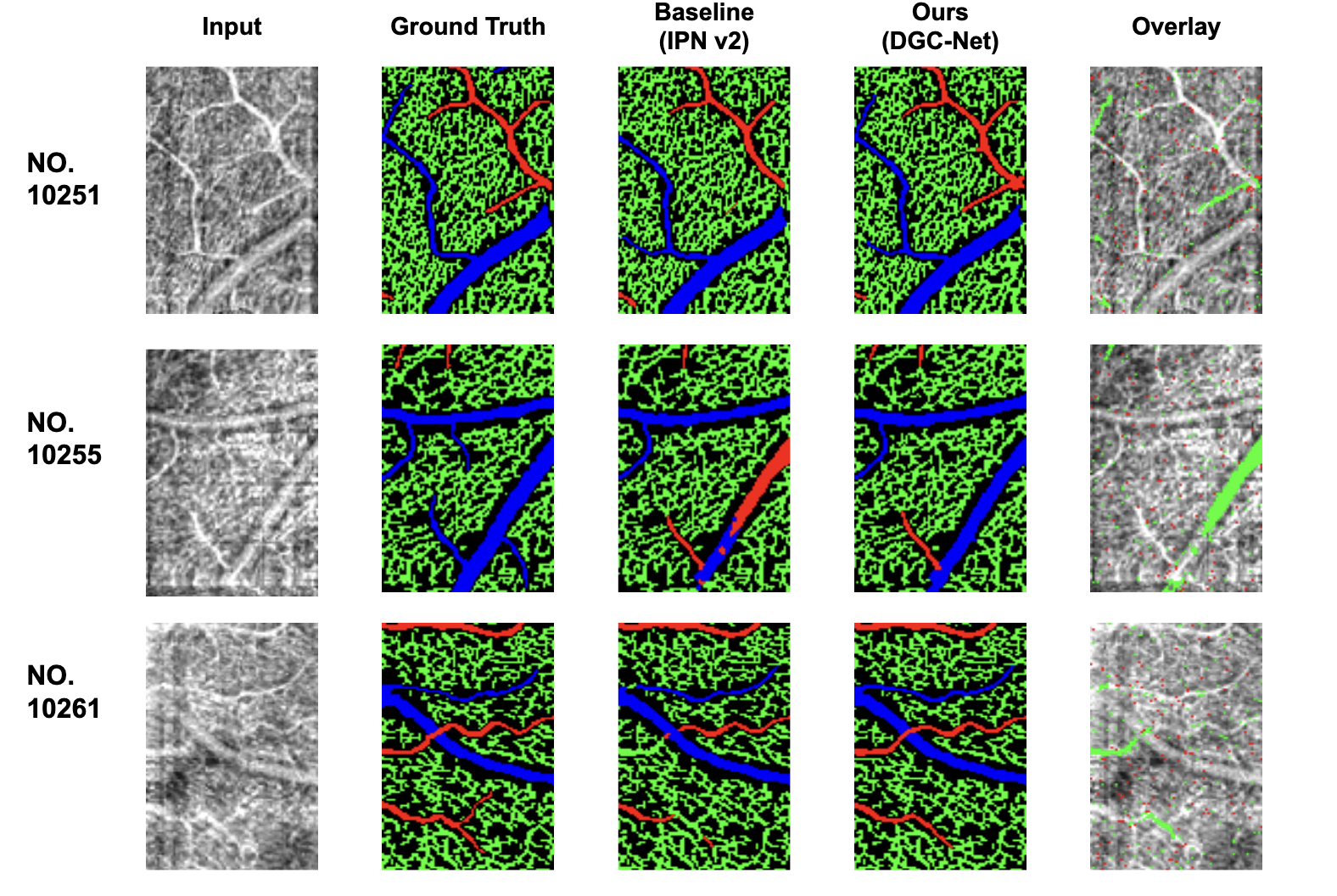} 
\vspace{-0.1in}
\caption{Qualitative comparison of topological disambiguation on the OCTA-6mm dataset. Columns from left to right: input projection, ground truth, IPNv2 baseline, DGC prediction, and the correction overlay. Classes are color-coded (Capillaries: green, Arteries: red, Veins: blue). In the overlay, green pixels denote macroscopic regions successfully recovered by DGC that were conflated by IPNv2, while red indicates residual microscopic boundary disagreements.}
\vspace{-0.05in}
\label{fig:qualitative}
\end{figure}

Figure~\ref{fig:qualitative} compares the IPNv2 baseline with DGC-Net at complex structural intersections, using zoomed patches from representative test volumes. The dominant failure mode of projection networks, which we term \emph{topological conflation}, appears at regions of macroscopic vascular overlap: in sample 10255, IPNv2 misclassifies a major vein as an artery, a direct consequence of collapsing vessels that occupy distinct depth strata into indistinguishable 2D intensities. By encoding continuous axial descriptors, the DM modules let DGC-Net separate the overlapping structures and recover the correct topology. Similar disambiguation is visible at the vascular junctions in other samples.

The correction overlays localize these gains: DGC-Net's recoveries (green) concentrate at artery--vein intersections, while residual disagreements (blue) are confined to fine capillary boundaries. This pattern is consistent with the robust objective suppressing noisy capillary labels while preserving the network's capacity to untangle macro-vascular geometry.

\subsection{Generalizability on Dermatological OCTA}

To test cross-domain generalization, we evaluate DGC-Net on the DERMA-OCTA dataset~\cite{Rotunno2025DERMAOCTA} for binary skin vessel segmentation. The original benchmark reports only Dice for a 2D U-Net on Average Intensity Projections (AIP); we re-implement this baseline under our unified protocol for a controlled comparison across both IoU and Dice. The comparative results are presented in Table~\ref{tab:derma}.

\begin{table}[t]
\centering
\setlength{\tabcolsep}{3pt}
\renewcommand{\arraystretch}{0.95}
\resizebox{\linewidth}{!}{
\begin{tabular}{lccc}
\hline
Method & Input Modality & IoU & Dice \\
\hline
\multicolumn{4}{c}{\textit{2D Methods}} \\
2D UNet (Re-impl.) & 2D AIP & $0.7115{\pm}0.0120$ & $0.8322{\pm}0.0144$ \\
\hline
\multicolumn{4}{c}{\textit{3D-to-2D Methods}} \\
IPNv2~\cite{octa500} & 3D Volume & $0.7451{\pm}0.0028$ & $0.8520{\pm}0.0024$ \\
H2C-Net~\cite{h2cnet} & 3D Volume & $0.7465{\pm}0.0034$ & $0.8517{\pm}0.0023$ \\
\textbf{DGC-Net (ours)} & 3D Volume & $\mathbf{0.7597{\pm}0.0017}$ & $\mathbf{0.8597{\pm}0.0016}$ \\
\hline
\end{tabular}
}
\vspace{-0.05in}
\caption{Cross-domain evaluation on the DERMA-OCTA dataset for binary skin vessel segmentation. Results are mean $\pm$ standard deviation over three independent training runs. DGC-Net generalizes to dermatological vasculature, outperforming both 2D and 3D-to-2D baselines with non-overlapping error bars.}
\vspace{-0.05in}
\label{tab:derma}
\end{table}

As shown in Table~\ref{tab:derma}, DGC-Net achieves the best cross-domain performance, recording $0.7597$ IoU and $0.8597$ Dice. Against the two 3D-to-2D baselines trained under identical conditions, DGC-Net improves IoU by $1.46$ pp over IPNv2 ($0.7451$) and $1.32$ pp over H2C-Net ($0.7465$), with non-overlapping standard-deviation bands in both cases. Notably, IPNv2 and H2C-Net are comparable on this dataset ($0.7451$ vs.\ $0.7465$), despite H2C-Net's substantially larger capacity---indicating that on out-of-domain dermatological vasculature, simply increasing decoder capacity does not help, whereas explicit depth-geometric conditioning does.

\looseness=-1 The gain is anatomically meaningful: dermatological microvasculature forms dense, chaotic tangles rather than planar beds. That DGC-Net’s advantage persists here indicates DM and HDM are not overfit to retinal topology and transfer to novel tissue types.


\section{Conclusion}
In this paper, we presented DGC-Net, a depth-geometrically conditioned framework for projection-supervised OCTA segmentation. By combining local Depth-Moment Conditioning with Holistic Depth-Geometry Modulation, DGC-Net preserves complementary axial localization, dispersion, and global layout cues during 3D-to-2D projection. Experiments on OCTA-500 and DERMA-OCTA demonstrate consistent improvements over strong 2D and 3D-to-2D baselines, while ablations and visualizations verify the contributions of the proposed components. Future work will investigate richer depth representations and projection-consistent learning under 2D supervision.
\section*{Acknowledgements}
This work was supported in part by Portuguese National Funds, through AICEP, E. P. E. and in part by the Innovation and Digital Transition Program under Grant COMPETE2030, and in part by Project ATLAS–Trusted Autonomous Navigation, with Funding Reference 01/RPA/2022-C679908640-00009887.
{
    \small
    \bibliographystyle{ieeenat_fullname}
    \bibliography{main}
}

\end{document}